\documentclass[10pt,twocolumn,letterpaper]{article}

\usepackage{iccv,times,epsfig,graphicx,amsmath,amssymb}
\usepackage{times,graphicx,amsmath,amssymb,booktabs,tabulary,multirow,overpic,xcolor,bbm,amssymb}
\usepackage{subcaption}

\newcolumntype{x}[1]{>{\centering\arraybackslash}p{#1pt}}

\newlength\savewidth

\makeatletter\renewcommand\paragraph{\@startsection{paragraph}{4}{\z@}
  {.5em \@plus1ex \@minus.2ex}{-.5em}{\normalfont\normalsize\bfseries}}\makeatother

\usepackage[square,numbers]{natbib}
\usepackage[pagebackref=true,breaklinks=true,letterpaper=true,colorlinks,bookmarks=false]{hyperref}

\iccvfinalcopy 

\def\iccvPaperID{45}

\ificcvfinal\fi
\begin{document}

\title{Class-Based Styling: Real-time Localized Style Transfer with Semantic Segmentation}

\author{Lironne Kurzman \\
University of British Columbia\\
{\tt\small lironkurz@gmail.com}
\and
David Vazquez \\
Element AI\\
{\tt\small dvazquez@elementai.com}
\and
Issam Laradji \\
Element AI; University of British Columbia\\
{\tt\small issam.laradji@gmail.com}
}

\maketitle
\begin{abstract}
We propose a Class-Based Styling method (CBS) that can map different styles for different object classes in real-time. CBS achieves real-time performance by carrying out two steps simultaneously. While a semantic segmentation method is used to obtain the mask of each object class in a video frame, a styling method is used to style that frame globally. Then an object class can be styled by combining the segmentation mask and the styled image. The user can also select multiple styles so that different object classes can have different styles in a single frame.
For semantic segmentation, we leverage DABNet that achieves high accuracy, yet only has 0.76 million parameters and runs at 104 FPS. For the style transfer step, we use the popular real-time method proposed by \citet{johnson2016perceptual}. We evaluated CBS on a video of the CityScapes dataset and observed high-quality localized style transfer results for different object classes and real-time performance. The code is available at \url{https://github.com/IssamLaradji/CBStyling}.
\end{abstract}


\section{Introduction}
\label{sec:intro}
One important aspect of style transfer is that it allows us to create new, beautiful artistic works. Style transfer maps the style of an image onto a target image, while maintaining the content of that target image. \citet{Jing2017NeuralST} conducted a review of neural style transfer methods, and evaluated their different applications. While style transfer methods have existed for over a decade, only recently did new methods emerge that leverage the powerful representation of deep learning. Perhaps \citet{gatys2015neural} proposed the first such style method that proved to be seminal work. The reason is that it enabled high-quality style creation by simply combining a style image and a content image. Unfortunately, this method is too slow to be used in real-time.

\begin{figure}[t]
  \centering
    \includegraphics[width=0.5\textwidth]{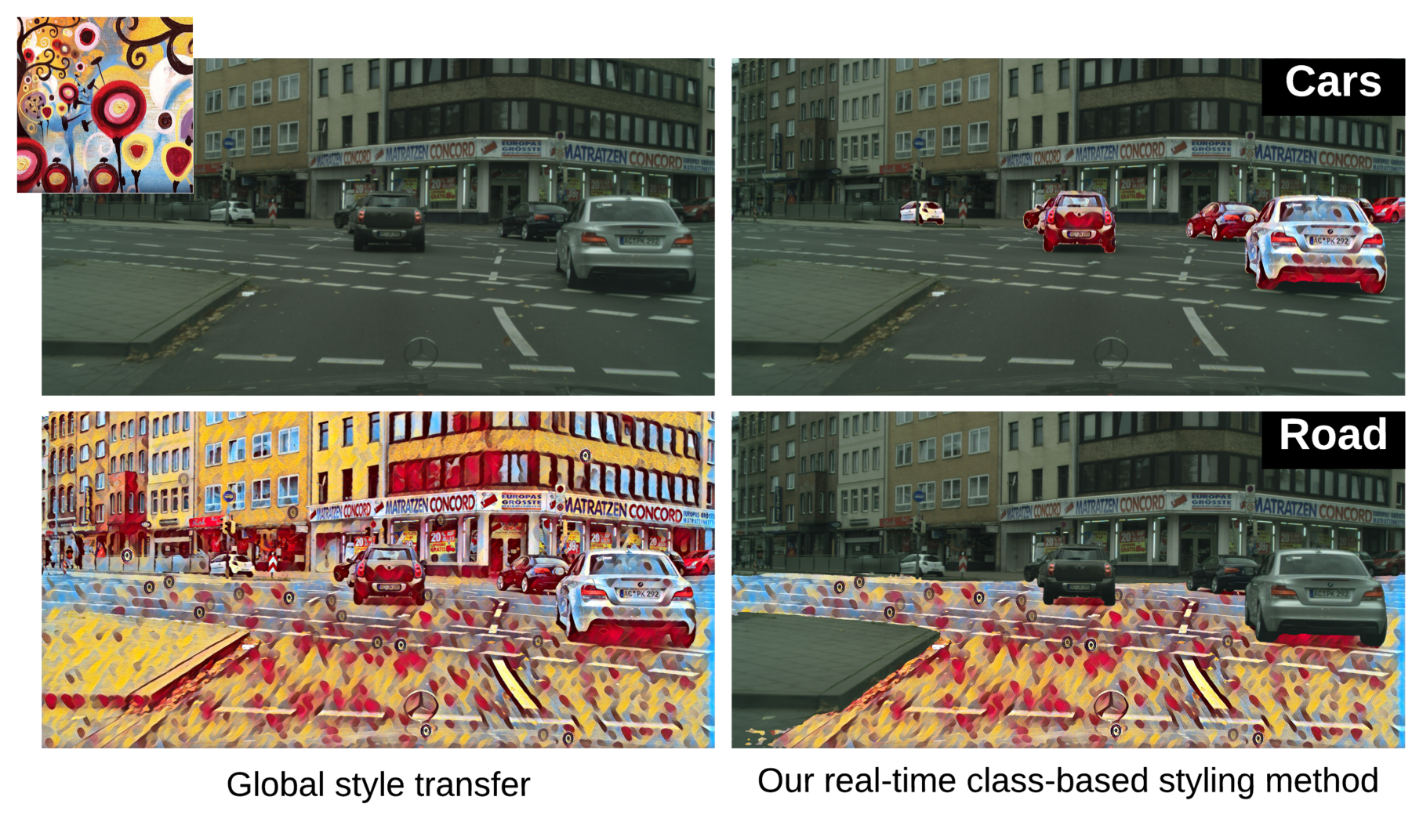}
  \caption{(Left) A global style transfer method from \citet{johnson2016perceptual} was applied to a CityScapes image. (Right) CBS (ours) was used to style {\it cars} and {\it road}.}
  \label{fig:main_figure}
\end{figure}

Later, another seminal work emerged that can run in real-time~\cite{johnson2016perceptual}. It differs from the method proposed in \citet{gatys2015neural} by using a deep network that performs a single feed-forward at inference time, instead of optimizing for the style and content at inference time. Namely, the method by \citet{johnson2016perceptual} directly transforms an input image to a styled image, which makes it fast enough to run in real-time. 

One limitation of \citet{johnson2016perceptual}'s method is that it transfers the style to the image globally. In fact, most methods have this limitation~\cite{jing2017neural}. \citet{castillo2017zorn} addressed this by proposing a localized style transfer method. It achieves this with the help of a Mask R-CNN~\cite{He2017MaskRCNN} to obtain object instances of the image. Then, it uses \citet{gatys2015neural} method to style a specific object instance that is specified by the user. Further, \citet{castillo2017zorn} proposed an MRF-based loss to smooth the boundary pixels of the stylized objects. Unfortunately, their method cannot run in real-time. This is because \citet{gatys2015neural} method, Mask R-CNN, and the MRF-based loss all have slow inference time.

\begin{figure*}[t]
  \centering
    \includegraphics[width=1.0\textwidth]{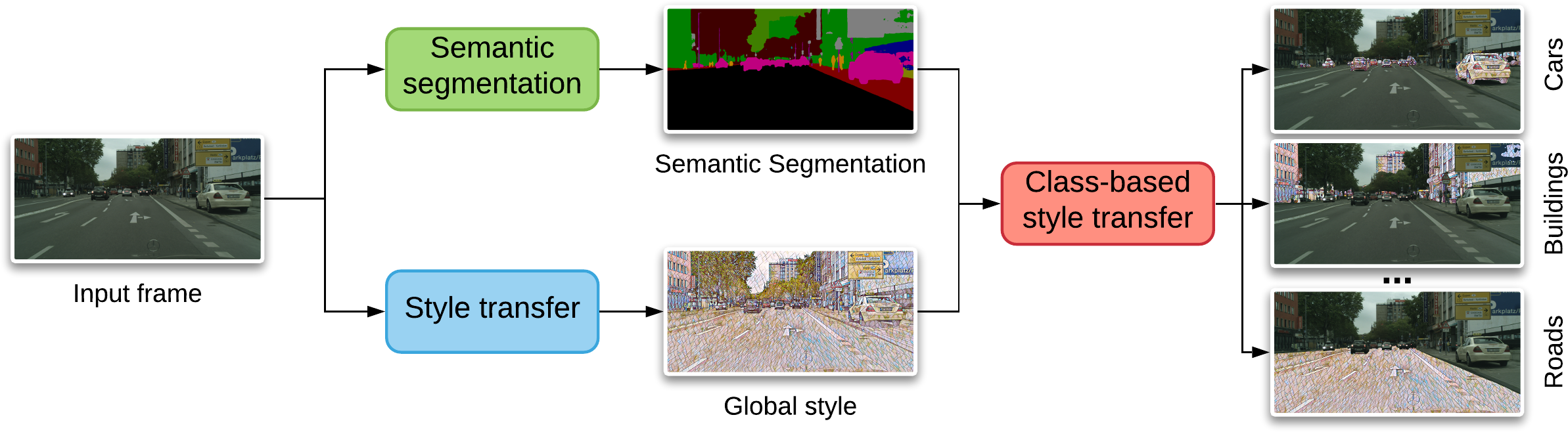}
  \caption{CBS pipeline. First, CBS takes a video frame as an input image and performs two operations in parallel: semantic segmentation and global style transfer. Then, the segmentation mask is combined with the styled image to style only the object classes of interest.}
  \label{fig:method}
\end{figure*}

In this work, we propose CBS, a Class-Based Styling method that performs localized style transfer in real-time. Given an input image, it extracts masks for different object classes using a pretrained DABNet~\cite{li2019dabnet}, a real-time semantic segmentation method (it runs at 104 fps for CityScapes~\cite{Cordts2016Cityscapes}). In parallel, it uses \citet{johnson2016perceptual} real-time style transfer method to map a style on that input image. 

CBS then applies the style only on those extracted masks. In this case, the user specifies which class the style should be applied to. For example, Figure~\ref{fig:main_figure} shows the style being applied to cars and road. Experimental results show that these two steps are fast enough to be real-time. Thus, we summarize our contributions as follows: (1) we propose a novel framework performing real-time localized style transfer, and (2) show that CBS achieves high-quality stylized images with real-time performance.

\section{Related Work}
\label{sec:related_work}

\paragraph{Style Transfer}
It was \citet{gatys2015neural} who first introduced the use of deep convolutional neural networks (CNNs) in whole-image styling, by minimizing both the content and, style-loss simultaneously. Despite producing beautiful results it wasn't very applicable to real-time videos due to the relatively slow inference speed. When \citet{johnson2016perceptual} extended their work using a feed-forward network, inference became substantially quicker and video style transfer became possible.

\paragraph{Real-time Semantic Segmentation}
Real-time semantic segmentation is a trade-off between accuracy and inference speed. Many of the frameworks available for a single-image semantic segmentation would fare poorly in real-time due to their slow inference. \citet{li2019dabnet} proposed a Depth-wise Asymmetric Bottleneck module to address the speed of semantic segmentation. They extract local and contextual features jointly, improving the speed without compromising accuracy. 

\paragraph{Localized Real-time Style Transfer}
Neural Style Transfer (NST) has been applied to specific regions of still images~\footnote{Found here: http://cs231n.stanford.edu/reports/2017/pdfs/416.pdf} through masking the objects of the target image, and modifying the original loss function to consider only the pixels of interest.
\citet{castillo2017zorn} have succeeded in creating images with partial styling as well. Their model classified pixels to foreground or background and applied the style to the pixels corresponding to objects selected by the user. They used Markov random fields (MRFs) to deal with the boundaries of the styled region in order smoothing out the outlier pixels as a way to anti-alias. However, applying their method does not work in real-time as they use \citet{gatys2015neural} method for the style transfer, which is slow.

\section{Proposed Method}
\label{sec:method}
We present CBS, a Class-Based Styling method. that can perform style transfer to specific object classes in real-time. To style an object class $c$ on an image $I$, CBS follows these steps.

First, using a fast segmentation network~\cite{li2019dabnet}, CBS extracts a binary mask $R_c$ for object class $c$. $R_c$ has the same height and width as $I$ with entries $1$ for the object belonging to class $c$ and $0$ otherwise. Note that the segmentation network is pretrained on segmenting object class $c$. Simultaneously, CBS transforms $I$ to a styled image $T_S$ using a fast styling method (we used the method by~\citet{johnson2016perceptual}).  

Then, using the binary mask $R_c$, CBS extracts the background $I_b$ from the unstyled image $I$ and extracts the foreground $T_f$ from the styled image $T_S$.
Finally, CBS adds $T_f$ and $I_b$ to obtain a target image $U$ that contains an unstyled background and a styled object class $c$ (Figure~\ref{fig:results}).

Figure~\ref{fig:method} provides an illustration of this pipeline. While this pipeline is similar to that proposed in \citet{castillo2017zorn}, theirs is slow due to the requirement of \citet{gatys2015neural} method, Mask R-CNN, and MRF blending. In contrast, our proposed framework can run in real-time with 16 fps. Below we explain CBS's components in detail.

\subsection{Fast Styling Method}
The goal is to generate a stylized image with the same content as input image $I$ in the style of an image $S$ in real-time. Thus, we use the fast styling method (FSM) proposed by \citet{johnson2016perceptual}. It trains a fully convolutional neural network (FCN) that learns to generate an image $T_S=M(I)$ with the content of the input image $I$ and style $S$. As a result, no optimization is required at inference time. Rather, the FCN performs a single forward pass for an image $I$ to get the stylized image. This leads to a high FPS rate when this method is applied on video. At training time, FSM optimizes the following loss function,
\begin{equation}
\begin{split}
\mathcal{L}(I) &= \frac{1}{C_2 H_2 H_2} \underbrace{||F_2(M(I)) - F_2(I)||_2^2}_{\text{content loss}} \\
&+ \underbrace{\sum_{l}^L \frac{1}{C_l}||G(M(I), l) - G(S, l)||_F^2}_{\text{style loss}}\\
\underbrace{G_{i,j}(X, l)}_{\text{gram matrix}} &= \frac{1}{H_l W_l} \sum_{h}^{H_l} \sum_{w}^{W_l} F_{l_{(h,w,i)}}(X) \cdot F_{l_{(h,w,j)}}(X),
\end{split}
\end{equation}
where $F_l(\cdot)$ is a feature extraction network (typically chosen as VGG16~\cite{Simonyan2014VGG}) that outputs a feature map at the $l$-th level (VGG16 outputs 4 levels of feature maps). $C_l \times H_l \times W_l$  is the dimension of the level $l$ feature map.

The content loss encourages FSM to preserve the spatial structure without preserving the color, texture and exact shape (these 3 properties define the style of the image). It can be considered as a feature reconstruction loss defined as the L2 norm between the feature representations of the input image and the generated image.

The style loss is there to preserve the style given by the style image $S$ which includes color and texture. It is computed as the uncentered covariance between the extracted features of an image, which captures the relationship between feature pairs. This is defined as the Gram matrix which evaluates which feature channels tend to activate together. Since this loss does not affect the spatial structure of the input image, this mainly encourages the input image to undergo a style change. On the other hand, the transformed input image keeps its spatial structure with the help of the content loss.

\subsection{Fast Segmentation Network}
The goal of the semantic segmentation component is to generate binary masks that represent the regions containing the object classes $C$ of interest. We use DABNet~\cite{li2019dabnet}, which is a recent semantic segmentation method that achieves real-time performance by combining depth-wise separable convolution and dilated convolution. As a result, it can extract dense features under a shallow network which allows many operations to run in parallel rather than sequentially. To train DABNet, we first use a labeled dataset with images $X_n$ and their corresponding one-hot encoded ground truth $Y_n$. The ground truth has the object classes that we wish to style. Next, we train DABNet to output the right prediction mask $S(X_n)$ by minimizing the following cross-entropy loss, 

\begin{equation}
\mathcal{L}_{ce} - \sum_{h,w}\sum_{c\in C} \log{(S(X_n)^{(h,w,c)})}.
\end{equation}

Finally, the network outputs the segmentation regions for object class $c$ as a binary matrix $R_c$.
Note that our framework is amenable to other real-time semantic segmentation methods.

\subsection{Styling the Object classes}
The final step of CBS is to generate an image $U$ that has stylized object classes with an unstylized background. For an object class $c$ and style $S$, the foreground mask (obtained by DABNet) is extracted for the stylized image (obtained by FSM) and then added to the background image. This is accomplished with the following element-wise product ($*$) and addition,
\begin{equation}
    U(c,S) = (R_c * T_S) +  (\mathbf{1} - R_c) * I.
\end{equation}


\begin{figure*}[t]
  \centering
  \includegraphics[width=1.0\textwidth]{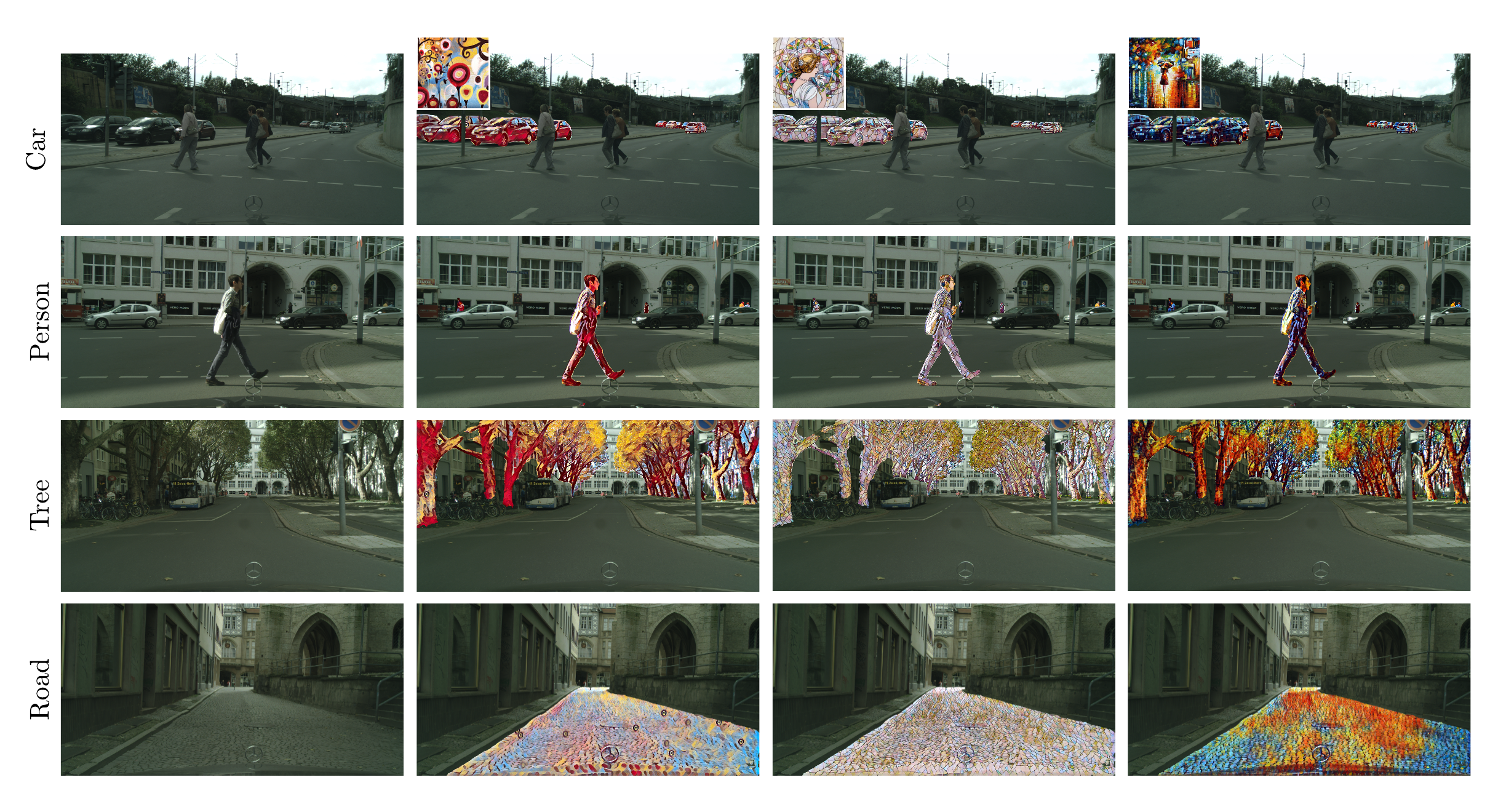}
  \caption{Qualitative results showing 3 different styles being applied to 4 different object classes.}
  \label{fig:results}
\end{figure*}

\section{Experiments} 
\label{sec:experiments}
We perform experiments to evaluate the efficacy of our method in two main aspects. Namely, the quality of the generated styles for different object classes; and the speed of CBS for styling video frames.

In Figure~\ref{fig:results} we show qualitative examples for a variety of style and content images. We see that the colors and textures of the style images successfully transferred to the object classes. In addition, the extracted segmentation masks correctly capture the objects of interest.

In our next benchmark, we ran CBS on Tesla P100 GPUs. Note that CBS was implemented in Pytorch~\cite{paszke2017pytorch}. For the Cityscapes~\cite{Cordts2016Cityscapes} $1024 \times 2048$ frames, CBS is able to process them at a speed of 16 FPS or 60.83 ms / image, making it feasible to run localized style transfer in real-time. Further, DABNet achieves 70.1\% Mean IoU on the CityScapes test set allowing it to achieve quality segmentation, which, in turn, makes the localized style transfer look more appealing. In contrast, \citet{castillo2017zorn} would take at least 15 seconds for a single image due to the \citet{gatys2015neural} method required at inference time. This makes it unsuitable for real-time style transfer. Thus, CBS can be used in more applications where artistic work requires real-time performance.

\section{Conclusion}
\label{sec:conclusion}
Our model achieves real-time localized style transfer by simultaneously segmenting and styling each frame. We anticipate that stylizing each class separately in real time can pave the way to more intricate videos, appealing works of arts, or in advertisements using product styling. For future work, we plan to extend this method for 3D scenes where localized style transfer can be applied to 3D objects.

{\small
\bibliography{paper}

\begin{thebibliography}{10}
\providecommand{\natexlab}[1]{#1}
\providecommand{\url}[1]{\texttt{#1}}
\expandafter\ifx\csname urlstyle\endcsname\relax
  \providecommand{\doi}[1]{doi: #1}\else
  \providecommand{\doi}{doi: \begingroup \urlstyle{rm}\Url}\fi

\bibitem[Castillo et~al.(2017)Castillo, De, Han, Singh, Yadav, and
  Goldstein]{castillo2017zorn}
C.~Castillo, S.~De, X.~Han, B.~Singh, A.~K. Yadav, and T.~Goldstein.
\newblock Son of zorn's lemma: Targeted style transfer using instance-aware
  semantic segmentation.
\newblock In \emph{ICASSP}, 2017.

\bibitem[Cordts et~al.(2016)Cordts, Omran, Ramos, Rehfeld, Enzweiler, Benenson,
  Franke, Roth, and Schiele]{Cordts2016Cityscapes}
M.~Cordts, M.~Omran, S.~Ramos, T.~Rehfeld, M.~Enzweiler, R.~Benenson,
  U.~Franke, S.~Roth, and B.~Schiele.
\newblock The cityscapes dataset for semantic urban scene understanding.
\newblock \emph{CVPR}, 2016.

\bibitem[Gatys et~al.(2015)Gatys, Ecker, and Bethge]{gatys2015neural}
L.~A. Gatys, A.~S. Ecker, and M.~Bethge.
\newblock A neural algorithm of artistic style.
\newblock \emph{arXiv:1508.06576}, 2015.

\bibitem[He et~al.(2017)He, Gkioxari, Doll{\'a}r, and Girshick]{He2017MaskRCNN}
K.~He, G.~Gkioxari, P.~Doll{\'a}r, and R.~B. Girshick.
\newblock Mask r-cnn.
\newblock \emph{ICCV}, 2017.

\bibitem[Jing et~al.(2017{\natexlab{a}})Jing, Yang, Feng, Ye, and
  Song]{Jing2017NeuralST}
Y.~Jing, Y.~Yang, Z.~Feng, J.~Ye, and M.~Song.
\newblock Neural style transfer: A review.
\newblock \emph{T-VCG}, 2017{\natexlab{a}}.

\bibitem[Jing et~al.(2017{\natexlab{b}})Jing, Yang, Feng, Ye, Yu, and
  Song]{jing2017neural}
Y.~Jing, Y.~Yang, Z.~Feng, J.~Ye, Y.~Yu, and M.~Song.
\newblock Neural style transfer: A review.
\newblock \emph{arXiv:1705.04058}, 2017{\natexlab{b}}.

\bibitem[Johnson et~al.(2016)Johnson, Alahi, and
  Fei-Fei]{johnson2016perceptual}
J.~Johnson, A.~Alahi, and L.~Fei-Fei.
\newblock Perceptual losses for real-time style transfer and super-resolution.
\newblock In \emph{ECCV}, 2016.

\bibitem[Li et~al.(2019)Li, Yun, Kim, and Kim]{li2019dabnet}
G.~Li, I.~Yun, J.~Kim, and J.~Kim.
\newblock Dabnet: Depth-wise asymmetric bottleneck for real-time semantic
  segmentation.
\newblock \emph{arXiv:1907.11357}, 2019.

\bibitem[Paszke et~al.(2017)Paszke, Gross, Chintala, Chanan, Yang, DeVito, Lin,
  Desmaison, Antiga, and Lerer]{paszke2017pytorch}
A.~Paszke, S.~Gross, S.~Chintala, G.~Chanan, E.~Yang, Z.~DeVito, Z.~Lin,
  A.~Desmaison, L.~Antiga, and A.~Lerer.
\newblock Automatic differentiation in {PyTorch}.
\newblock In \emph{NIPS Autodiff Workshop}, 2017.

\bibitem[Simonyan and Zisserman(2014)]{Simonyan2014VGG}
K.~Simonyan and A.~Zisserman.
\newblock Very deep convolutional networks for large-scale image recognition.
\newblock \emph{CoRR}, abs/1409.1556, 2014.

\end{thebibliography}
\bibliographystyle{plain}
}

\end{document}